\documentclass[letterpaper, 10 pt, conference]{ieeeconf}  

\IEEEoverridecommandlockouts                              

\usepackage{graphics} 
\usepackage{epsfig} 
\usepackage{mathptmx} 
\usepackage{times} 
\usepackage{amsmath} 
\usepackage{amssymb}  
\usepackage{booktabs}
\usepackage{pifont}
\usepackage{tabularx}
\usepackage[table]{xcolor}
\usepackage{verbatim}
\usepackage{multirow}
\usepackage{hyperref}
\usepackage{cleveref}

\newcommand{\cmark}{\textcolor{green!60!black}{\ding{51}}}
\newcommand{\xmarkr}{\textcolor{red!70!black}{\ding{55}}}

\title{\LARGE \bf
Unified Map Prior Encoder for Mapping and Planning
}
\author{Zongzheng Zhang$^{1, 2*}$, Sizhe Zou$^{1*}$, Guantian Zheng$^{1*}$, Zhenxin Zhu$^{1}$, Yu Gao$^{2}$,\\ Guoxuan Chi$^{1}$, Shuo Wang$^{2}$, Yuwen Heng$^{2}$, Zhigang Sun$^{2}$, Yiru Wang$^{2}$, \\Hao Sun$^{2}$, Chao Ma$^{3}$, Zhen Li$^{4}$, Anqing Jiang$^{2\dagger}$, Hao Zhao$^{1\dagger}$%
\thanks{%
    $^{1}$Institute for AI Industry Research (AIR), Tsinghua University.
    $^{2}$Bosch Corporate Research, China.
    $^{3}$Shanghai Jiao Tong University.
    $^{4}$Chinese University of Hong Kong, Shenzhen.
    $^{*}$Equal contribution. $^{\dagger}$Equal advising.%
}%
}

\begin{document}

\maketitle

\begin{abstract}

Online mapping and end-to-end (E2E) planning in autonomous driving are still largely sensor-centric, leaving rich map priors (HD/SD vector maps, rasterized SD maps, and satellite imagery) underused due to heterogeneity, pose drift, and inconsistent availability at test time. We present \emph{UMPE}, a Unified Map Prior Encoder that can ingest any subset of four priors and fuse them with BEV features for both mapping and planning. \emph{UMPE} has two branches. The vector encoder pre-aligns HD/SD polylines with a frame-wise SE(2) correction, encodes points via multi-frequency sinusoidal features, and produces polyline tokens with confidence scores. BEV queries then apply cross-attention with confidence bias, followed by normalized channel-wise gating to avoid length imbalance and to softly down-weight uncertain sources. The raster encoder shares a ResNet-18 backbone conditioned by FiLM (scaling/shift at every stage), performs SE(2) micro-alignment, and injects priors through zero-initialized residual fusion so the network starts from a do-no-harm baseline and learns to add only useful prior evidence. A vector-then-raster fusion order reflects the inductive bias of “geometry first, appearance second.” On nuScenes mapping, \emph{UMPE} lifts MapTRv2 from $61.5 \rightarrow 67.4$ mAP ($+5.9$) and MapQR from $66.4 \rightarrow 71.7$ mAP ($+5.3$). On Argoverse2, \emph{UMPE} adds $+4.1$ mAP over strong baselines. \emph{UMPE} is compositional: when trained with all priors, it outperforms single-prior models even when only one prior is available at test time, demonstrating powerset robustness. For E2E planning (VAD backbone, nuScenes), \emph{UMPE} reduces trajectory error from $0.72 \rightarrow 0.42$ m L2 (avg.\ $-0.30$ m) and collision rate from $0.22\% \rightarrow 0.12\%$ ($-0.10\%$), surpassing recent prior-injection methods. These results show that a unified, alignment-aware treatment of heterogeneous map priors yields better mapping and better planning. Code and dataset are released at \texttt{https://github.com/Ethan-Zheng136/UMPE}


\end{abstract}
\section{Introduction}

\begin{figure}
    \centering
    \includegraphics[width=0.95\linewidth]{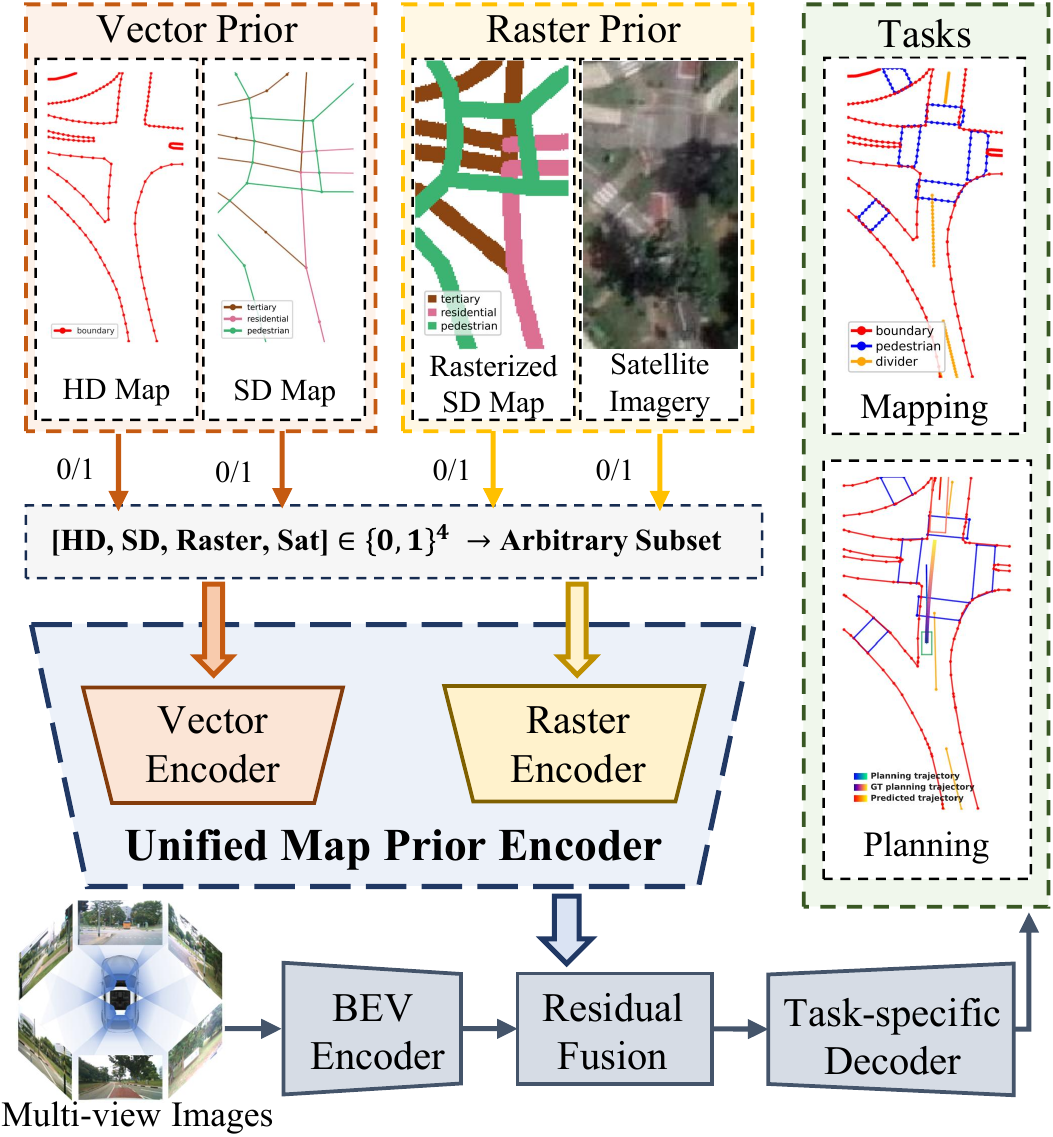}
    \caption{\textbf{Unified Map Prior Encoder (\emph{UMPE})}. \emph{UMPE} ingests an arbitrary subset of four map priors—vector (HD/SD vectorized maps) and raster (rasterized SD map, satellite imagery), and processes them via a vector encoder and a raster encoder. The resulting priors are fused with BEV features, supporting both online HD mapping and end-to-end planning tasks.}
    \label{fig:teaser}
    \vspace{-20pt}
\end{figure}

Most prior works inject one kind of map prior~\cite{smerf2024, ras_sdmap_gnn2024, sdtagnet2025, SatforHDmap_2024} or a fixed pair~\cite{smart2025, SEPT2025, SATP2025CVPR, unified_vector_prior} into sensor-centric autonomous driving pipelines, which leaves heterogeneous sources hard to combine when availability changes at test time (Tab.~\ref{Tab:comparison-map-priors}). In contrast, we introduce a unified setting (Fig.~\ref{fig:teaser}) where a single encoder can ingest any subset of four complementary map priors—HD/SD vector maps and raster priors (rasterized SD maps, satellite imagery)—and fuse them with BEV features for both online mapping and end-to-end planning. This “powerset” formulation is, to our knowledge, the first to treat map priors as interchangeable signals that can be turned on/off without retraining. Fig.~\ref{fig:teaser} visualizes this design: heterogeneous inputs enter two branches (vector and raster), are aligned and confidence-weighted, and are merged into a single BEV representation shared by mapping and planning heads.

Real-world deployments rarely enjoy a single perfect prior. HD vectors may exist in downtown but not in suburbs; SD map coverage is broad but coarse; satellite context is global yet misaligned; and rasterized SD provides topology hints when vectors are missing. Fig.~\ref{fig:teaser} makes this concrete: each prior is togglable (0/1), so our encoder can gracefully degrade—e.g., plan with only SD+rasterized SD when HD is absent, or tighten lane geometry with HD while satellite improves crosswalk texture. The performance also improves even when all priors are present: co-training across sources teaches the model to reconcile geometry (“vector first”) with appearance (“raster second”), yielding better BEV features for both mapping and planning.

Our vector encoder pre-aligns HD/SD polylines to the BEV frame via a small frame-wise SE(2) correction, encodes points with multi-frequency sinusoidal features, and produces polyline tokens with confidences. BEV queries then perform dual cross-attention to HD and SD separately to avoid softmax length imbalance, with an additive log-confidence bias that down-weights uncertain vectors. A presence-normalized, channel-wise gate mixes sources so that, when one prior is missing (the dashed 0/1 switches in Fig.~\ref{fig:teaser}), its channels do not suppress others. This path provides metrically precise lane geometry to downstream heads.

Fig.~\ref{fig:teaser}'s raster encoder shares a ResNet-18 backbone across satellite and rasterized SD inputs and conditions it with FiLM at every stage for source awareness. We estimate a lightweight SE(2) micro-alignment to correct residual pose/scale offsets to the BEV lattice. Priors are then injected through a zero-initialized residual pathway into the BEV tokens, with LayerNorm and a learnable scale, implementing a do-no-harm baseline that only adds evidence the task demands. A presence-normalized gate (as in the vector path) selects between satellite and rasterized SD features.

The two branches are composed in a vector-then-raster order that encodes an inductive bias—“geometry first, appearance second.” This sequencing preserves clean queries for vector attention and lets raster cues refine dense context afterward. To make the model robust to missing inputs, we introduce SourceDropout that randomly disables sources during training. Together, confidence-biased attention, zero-init residual fusion, and gated mixing yield a single encoder that generalizes across all prior combinations without per-subset retraining.

We validate the unified design in Fig.~\ref{fig:teaser} on two fronts. For online mapping, inserting \emph{UMPE} into strong BEV baselines (e.g., MapTRv2~\cite{Maptrv2} and MapQR~\cite{liu2024mapQR}) boosts mAP on nuScenes~\cite{2020nuscenes} and Argoverse 2~\cite{wilson2023argoverse}, with per-class gains matching intuition: vector priors enhance boundaries/dividers, while raster priors sharpen pedestrian crossings. For E2E planning, we plug \emph{UMPE} into a VAD-style backbone~\cite{jiang2023vad}, reducing average trajectory $\mathbf{L}_2$ and collision rate versus prior-injection methods. Ablations validate each design choice, and robustness tests show that a model trained with all priors still outperforms single-prior models even when only one prior is available at test time—the practical payoff of our proposed unified setting.

\begin{table}[t]
  \centering
   \caption{\small Priors used and downstream tasks across methods. Task: M--mapping; T--Topology; P--Planning}
  \resizebox{\linewidth}{!}{
  \begin{tabular}{l|c|cccc|c}
    \toprule
    \textbf{Method} & \textbf{Venues}& \textbf{HD map (vec)} & \textbf{SD map (vec)} & \textbf{Sat. Ima.} & \textbf{SD map (ras)} & \textbf{Task} \\
    \midrule
    SMERF~\cite{smerf2024} &ICRA 2024 &\xmarkr  & \cmark & \xmarkr & \xmarkr & M\&T \\
    SDmap-GNN~\cite{ras_sdmap_gnn2024}&IROS 2024 & \xmarkr & \xmarkr & \xmarkr & \cmark & M\&T \\
    SDTagNet~\cite{sdtagnet2025}& Arxiv 2025 & \xmarkr & \cmark & \xmarkr & \xmarkr & M \\
    SatforHDMap~\cite{SatforHDmap_2024}& ICRA 2024  & \xmarkr & \xmarkr & \cmark & \xmarkr & M \\
    \midrule
    SMART~\cite{smart2025}& ICRA 2025 & \xmarkr & \cmark & \cmark & \xmarkr & M\&T \\
    SEPT~\cite{SEPT2025}& RAL 2025 & \xmarkr & \cmark & \xmarkr & \cmark & M\&T \\
    SATP~\cite{SATP2025CVPR}& CVPR 2025 & \cmark & \cmark & \xmarkr & \xmarkr & P \\
    PriorDrive~\cite{unified_vector_prior}& Arxiv 2024 & \cmark & \cmark & \xmarkr & \xmarkr & M \\
    \midrule
    \rowcolor{gray!20}
    \textbf{UMPE (Ours)}& --- & \cmark & \cmark & \cmark & \cmark & M\&P \\
    \bottomrule
  \end{tabular}
  }
  \label{Tab:comparison-map-priors}
  \vspace{-10pt}
\end{table}

\section{Related Work}

\subsection{Online HD Mapping and Motion Forecasting}
Online HD mapping has evolved from BEV-based frameworks~\cite{Hdmapnet} to end-to-end vectorized prediction with set-structured queries~\cite{liu2023vectormapnet,maptr2022,Maptrv2}, later extended to temporal streaming and instance-consistent tracking~\cite{Streammapnet2024,2024maptracker} and hybrid raster-vector models~\cite{2024himap}. In parallel, detection and topology reasoning~\cite{zhang2025chameleon, li2025reusing} have been standardized by OpenLane-V2~\cite{openlanev2}, with graph-based~\cite{TopoNet}, lightweight MLP~\cite{topomlp} approaches and lane-segment perception~\cite{lanesegnet} enriching the task definition. 

Motion forecasting has similarly progressed from goal/intention-driven models~\cite{2024mtr++,gu2021densetnt,2022hivt,2023qcnet} toward tighter coupling with online maps, e.g., direct BEV feature attention~\cite{2024uncexxv} and explicit map-uncertainty modeling~\cite{gu2024unccvpr}. Yet most pipelines remain \emph{sensor-centric} (camera/LiDAR), leaving complementary map priors underexploited.

\subsection{Map Prior for Online Mapping}
Recent work increasingly augments online mapping with heterogeneous priors—prebuilt HD maps, standard-definition (SD) maps, satellite imagery, and neural radiance fields~\cite{yuan2024presight, xiong2023neural}—while tackling their misalignment with onboard perception and representation gaps across modalities. 

HD/SD maps provide vector road skeletons, which, when encoded and fused with onboard features, improve mapping and topology~\cite{smerf2024, sdtagnet2025, ras_sdmap_gnn2024}. Satellite imagery contributes global, long-range context with feature-level fusion and BEV-frame alignment~\cite{SatforHDmap_2024}. Beyond single sources, mixed priors yield further gains: HD+SD for far-seeing generation~\cite{pmapnet2024}, SD+satellite priors learned offline then plugged into topology heads~\cite{smart2025}, SD (vector)+rasterized SD via dual-branch fusion~\cite{SEPT2025}, and explicit HD--SD alignment beneficial to mapping and planning~\cite{SATP2025CVPR}. A unified vector prior encoder pushes toward \emph{map-type--agnostic} consumption by embedding SD/HD into a shared space~\cite{unified_vector_prior}.

Accordingly, we further propose a single unified map prior encoder that jointly learns from vectorized HD/SD, satellite, and rasterized SD priors with alignment-aware features.

\subsection{Map Prior for End-to-End Autonomous Driving}
E2E driving spans planning-oriented multi-task models~\cite{UniAD,jiang2023vad,bevplanner2024,2024hydramdp, zhang2025delving, gao2026uniuncer}, generative trajectory policies~\cite{2025diffusiondrive,2025goalflow, 2024genad}, world-model or sparse-token formulations~\cite{sun2024sparsedrive,2025epona}, and specialized designs with temporal memory or language guidance~\cite{2025MomAD,2025bridgeAD,2025orion}. 

Within this landscape, explicit priors have shown clear benefits for E2E planning: SATP~\cite{SATP2025CVPR} aligns SD–HD maps and improves closed-loop planning when the aligned priors are fed into VAD-style stacks; GaussianFusion~\cite{2025gaussianfusion} uses a gaussian-based multi-sensor fusion framework, offering a compact alternative to dense BEV features. Extending this direction, we present a unified encoder that integrates well-aligned priors into a single representation for E2E planning.


\begin{figure*}
    \centering
    \includegraphics[width=1.0\linewidth]{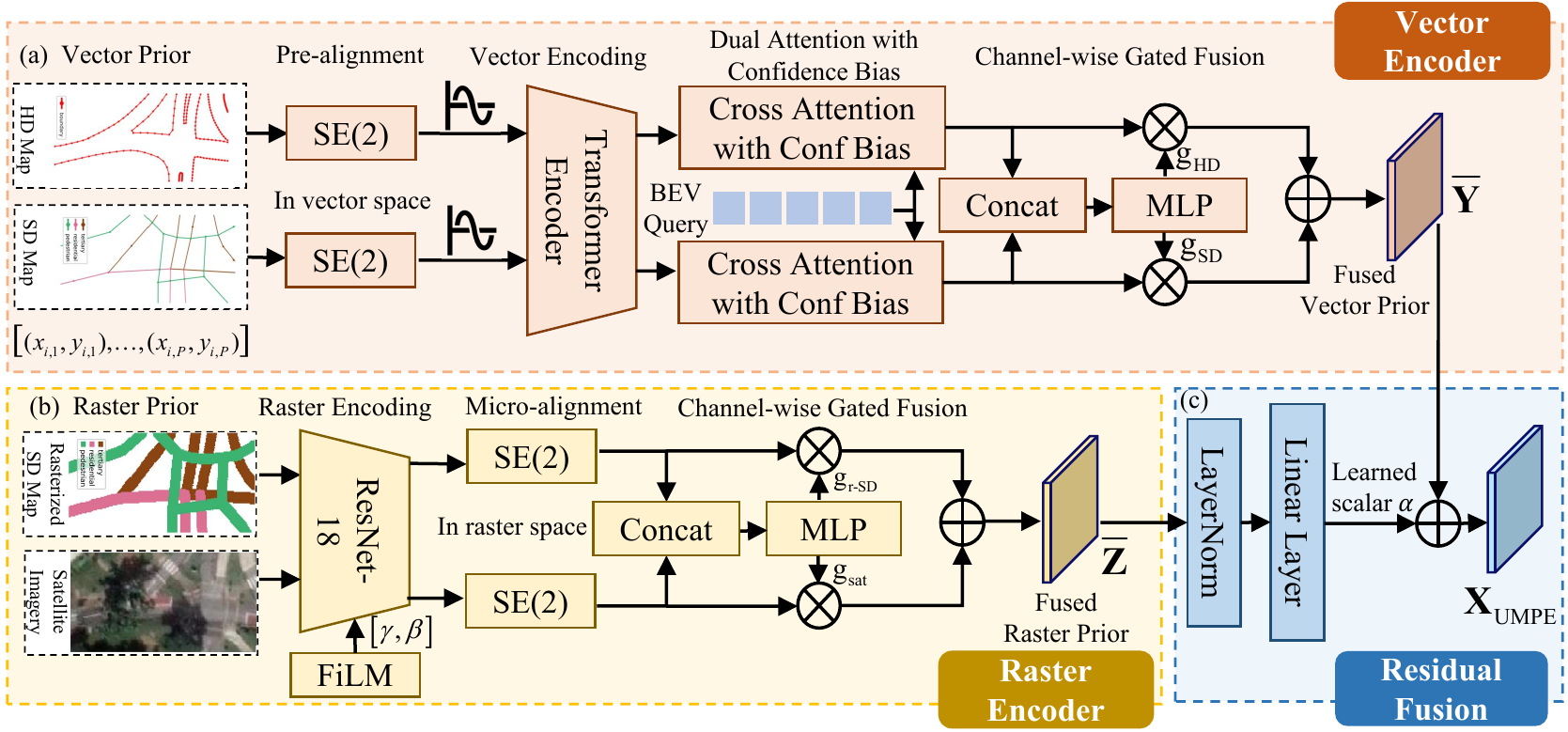}
    \caption{\textbf{Unified Map Prior Encoder (\emph{UMPE}) architecture.} (a) \textbf{Vector Encoder}: HD/SD polylines are $\mathrm{SE}(2)$ pre-aligned and encoded; BEV queries attend to each source with confidence-biased dual cross-attention. Presence-normalized, channel-wise gating mixes sources to produce fused vector tokens $\bar{\mathbf{Y}}$. (b) \textbf{Raster Encoder}: rasterized SD map and satellite imagery pass through a shared FiLM-conditioned ResNet, then undergo $\mathrm{SE}(2)$ micro-alignment in raster space; channel-wise gating yields fused raster tokens $\bar{\mathbf{Z}}$. (c) \textbf{Residual fusion}: $\bar{\mathbf{Y}}$ and $\bar{\mathbf{Z}}$ are inhected with a learned scalar $\alpha$, producing $\mathbf{X}_{\mathrm{UMPE}}$.}
    \vspace{-10pt}
    \label{fig:method}
\end{figure*}

\section{Method}
In this section, we first describe how four map priors  are obtained (Sec.~\ref{sec: priors preperation}). We then present the detailed architecture of the vector encoder (Sec.~\ref{Sec: vector encoder}) and the raster encoder (Sec.~\ref{Sec: raster encoder}). Finally, we show the unified fusion that integrates the two encoder outputs with BEV tokens (Sec.~\ref{sec: four priors fusion}), enabling \emph{UMPE} to handle any subset of priors.

\subsection{Map Priors Preparation}
\label{sec: priors preperation}
\textbf{Vectorized HD and SD Map.} 
We retrieve SD maps from OpenStreetMap~\cite{haklay2008openstreetmap}. For each frame, given the ego GPS and heading orientation, we query a local OSM region, project coordinates to a local Cartesian frame, apply a rigid transform to the ego frame (rotation by yaw and translation by ego position), and crop an ego-centric BEV window of $60{\rm m}\!\times\!30{\rm m}$, the same as the spatial extent covered by $\mathbf{F}_{\mathrm{BEV}}$. OSM roadways are parsed as polylines and annotated with eight one-hot classes: \texttt{highway}, \texttt{primary}, \texttt{secondary}, etc. The dataset-provided HD vectors undergo the same projection, rigid transform, and crop. Both SD and HD coordinates are expressed in the exact coordinate system of the $\mathbf{F}_{\mathrm{BEV}}$. To standardize density and batching, every polyline is uniformly resampled to the same number of points.

\textbf{Satellite Imagery.} We fetch satellite tiles via the Mapbox Raster Tiles API given the ego GPS. We compute the tile indices at a zoom level chosen to match the pixel–meter resolution of $\mathbf{F}_{\mathrm{BEV}}$. All tiles covering the $60{\rm m}\times30{\rm m}$ area around the ego are downloaded, then rotated by the ego yaw. The result is an RGB image $\mathbf{I}^\text{sat}\in\mathbb{R}^{H\times W\times 3}$ for every frame; $H$ and $W$ equal the BEV canvas used by the network.

\textbf{Rasterized SD Map.} Given the SD map, we render each of the eight SD categories with a fixed color. This produces an RGB raster $\mathbf{I}^\text{sd}\in\mathbb{R}^{H\times W\times 3}$ used as the rasterized SD prior.

\subsection{Vector Encoder for Vectorized HD/SD Map Priors}
\label{Sec: vector encoder}
Vector map priors provide exact lane geometry but arrive with small pose drift and a variable number of polylines that do not align to a fixed BEV grid. We therefore (i) correct coordinates by a small frame-wise \textbf{$\mathrm{SE}(2)$ motion}, (ii) encode each polyline as a fixed-width token using sinusoidal point features plus semantics (category and source one-hots), then apply a transformer to obtain tokens and confidence, and (iii) fuse into BEV by \textbf{dual cross-attention} with \textbf{confidence bias} and \textbf{presence normalized gating}, so that BEV queries selectively pull geometrically relevant, reliable vectors while softly suppressing uncertain or absent sources (Fig.~\ref{fig:method} (a)).

\textbf{Coordinate-level SE(2) Pre-alignment.}
To compensate for small pose drift between the visual BEV frame and the vector priors, we estimate a \emph{frame-wise} rigid correction for each available source $\text{src}\!\in\!\{\mathrm{HD},\mathrm{SD}\}$. 
For source $\text{src}$, the resampled polyline set is
$
\mathcal{P}^{\text{src}}=\{\mathbf{p}_i^{\text{src}}\}_{i=1}^{N_{\text{src}}},
\quad 
\mathbf{p}_i^{\text{src}}=\big[(x_{i,1},y_{i,1}),\ldots,(x_{i,P},y_{i,P})\big],\ P{=}11,
$
where $N_{\text{src}}$ is the number of polylines in the current frame.
We predict a small rigid motion \((\Delta{x},\Delta{y},\Delta{\theta})\) and correct every point:
\begin{equation}
    \tilde{\mathbf{p}_i} = \mathbf{R}(\Delta\theta)\mathbf{p}_i+\mathbf{T}, \ 
    \mathbf{R}(\Delta\theta) = \left[\begin{smallmatrix} 
    \cos\Delta\theta & -\sin\Delta\theta \\ 
    \sin\Delta\theta & \cos\Delta\theta 
    \end{smallmatrix}\right], \ 
    \mathbf{T} = [\Delta x, \Delta y]^\top.
\end{equation}
We regularize its magnitude
\(
\mathcal{L}_{\mathrm{se2}}^{\mathrm{vec}}
=
\lambda_t\,\|\mathbf{T}\|_2^2 \;+\;
\lambda_r(\Delta{\theta})^2
\)
to keep corrections small.

\textbf{Vector Encoding and Tokenization.}
Each corrected point $(\tilde x,\tilde y)$ is mapped by a multi-frequency sinusoidal embedding
$
\phi(\tilde x,\tilde y)=\big[\sin(\omega_k\tilde x),\cos(\omega_k\tilde x),\sin(\omega_k\tilde y),\cos(\omega_k\tilde y)\big],
$
with geometrically spaced \(\omega_k\). For polyline \(i\), we \emph{flatten} its \(P\) point encodings and concatenate a one-hot category \(\mathbf{e}^{\mathrm{cat}}_i\in\{0,1\}^{K_{\mathrm{cat}}}\) and one-hot source \(\mathbf{e}^{\mathrm{src}}_i\in\{0,1\}^{2}\):
$
\mathbf{z}_i
=
\Big[
\mathrm{vec}\big(\phi(\tilde{\mathbf{x}}_{i,1}),\ldots,\phi(\tilde{\mathbf{x}}_{i,P})\big)\;;\;
\mathbf{e}^{\mathrm{cat}}_i\;;\;
\mathbf{e}^{\mathrm{src}}_i
\Big].
$
We then apply a 6-layer transformer encoder:
$ \mathbf{T}_{\text{vec},i}^{\text{src}}=\mathrm{TrEnc}^{(6)}(\mathbf{z}_i),$
and predict a sigmoid confidence for each token
$U_i^{\text{src}}\in(0,1).$

\textbf{Dual Cross-Attention with Confidence Bias.}
Let BEV feature $\mathbf{F}_{\mathrm{BEV}}\!\equiv\!\mathbf{X}\in\mathbb{R}^{ B\times(HW)\times C}$ be the BEV tokens (queries), and let 
$\mathbf{T}_{\text{vec}}^{\text{src}}\in\mathbb{R}^{B\times N_{\text{src}}\times C}$ be the contextualized polyline tokens from source $\text{src}\!\in\!\{\mathrm{HD},\mathrm{SD}\}$.
We compute multi-head projections
$\mathbf{Q}=\mathbf{X}\mathbf{W}_Q,$
$\mathbf{K}^{\text{src}}=\mathbf{T}_{\text{vec}}^{\text{src}}\mathbf{W}_K,$
$\mathbf{V}^{\text{src}}=\mathbf{T}_{\text{vec}}^{\text{src}}\mathbf{W}_V,$
where $\mathbf{W}_Q,\mathbf{W}_K,\mathbf{W}_V\in\mathbb{R}^{C\times (h d)}$.
Let $\tilde{\mathbf{U}}^{\text{src}}=\mathrm{clamp}(\mathbf{U}^{\text{src}},\varepsilon,1)\in(0,1]^{B\times N_{\text{src}}}$ be the confidence with lower bound $\varepsilon$.
For each source, we fuse separately to avoid length imbalance in a single softmax:

\begin{equation}
    \mathbf{Y}^{\text{src}}
=\mathrm{softmax}\!\left(
\frac{\mathbf{Q}\,(\mathbf{K}^{\text{src}})^\top}{\sqrt{d}}
+\log \tilde{\mathbf{U}}^{\text{src}}
\right)\mathbf{V}^{\text{src}}
\quad\in\mathbb{R}^{B\times (HW)\times C},
\label{eq: cross attention}
\end{equation}
where $\log \tilde{\mathbf{U}}^{\text{src}}$ is broadcast along query positions and heads to form a $(B\times HW\times N_{\text{src}})$ bias matrix.
The additive $\log\tilde{\mathbf{U}}^{\text{src}}$ acts as a multiplicative prior inside the softmax, privileging reliable polylines while softly suppressing uncertain ones.

\textbf{Presence-normalized Channel-wise Gated Fusion.} 
After dual cross-attention, we obtain $\mathbf{Y}^{\mathrm{HD}}$, $\mathbf{Y}^{\mathrm{SD}}$; we concatenate along channels and pass through a lightweight network to produce logits $\mathbf{L}\in\mathbb{R}^{B\times(2C)}$. Splitting $\mathbf{L}$ into $\mathbf{L}_{\mathrm{HD}},\mathbf{L}_{\mathrm{SD}}\in\mathbb{R}^{B\times C}$, we compute per-channel, presence-normalized gates via a softmax across sources:
\begin{equation}
\begin{aligned}
\big[g_{\mathrm{HD}},\,g_{\mathrm{SD}}\big]
&= \mathrm{softmax}\!\Big(
\big[\mathbf{L}_{\mathrm{HD}}+\log(\pi_{\mathrm{HD}}{+}\varepsilon),\\
&\qquad\qquad\ \mathbf{L}_{\mathrm{SD}}+\log(\pi_{\mathrm{SD}}{+}\varepsilon)\big]
\Big).
\end{aligned}
\label{eq:channelwise_gates}
\end{equation}
where $\boldsymbol{\pi}=[\pi_{\mathrm{HD}},\pi_{\mathrm{SD}}]\in\{0,1\}^2$ is the source-presence mask for the current frame
($\pi_s{=}1$ if source $s$ is available, $0$ otherwise).
We then take the gated mix as the fused vector-prior BEV tokens:
\begin{equation}
    \bar{\mathbf{Y}}=g_{\mathrm{HD}}\odot \mathbf{Y}^{\mathrm{HD}}+g_{\mathrm{SD}}\odot \mathbf{Y}^{\mathrm{SD}}
\in\mathbb{R}^{B\times(HW)\times C}.
\label{eq:gate_fusion}
\end{equation}

\subsection{Raster Encoder for Satellite and Rasterized SD Priors}
\label{Sec: raster encoder}
We introduce a source-aware raster encoder that ingests two raster priors. The module has three stages: \textbf{source-aware encoding} (shared backbone with FiLM conditioning), \textbf{SE(2) micro-alignment} to correct residual pose/scale mismatches, and \textbf{gated fusion} to produce fused raster tokens \(\bar{\mathbf{Z}}\) (Fig.~\ref{fig:method}(b)).


\textbf{Source-aware Backbone with FiLM.}
Both sources $\mathbf{I}^{sat}, \mathbf{I}^{sd}\in\mathbb{R}^{H\times W\times 3}$ are processed by a shared ResNet-18 backbone~\cite{he2016deep} equipped with every-stage FiLM conditioning. Let $\mathbf{A}\in\mathbb{R}^{B\times C\times H\times W}$ be an intermediate activation and $c\in\mathbb{R}^{D}$ a learned source embedding (one-hot identity passed through an MLP). FiLM computes channel-wise affine parameters:
\begin{equation}
[\gamma,\beta]=\mathbf{W}c+\mathbf{b}\in\mathbb{R}^{2C},\qquad\\
\mathrm{FiLM}(\mathbf{A},c)=(1+\gamma)\odot \mathbf{A}+\beta,
\label{eq:film raster}
\end{equation}
broadcast over spatial dimensions. 
A $1\times 1$ projection followed by resizing produces source-aligned BEV feature:
\begin{equation}
\begin{aligned}
    \mathbf{F}^{\text{src}}_{\text{ras}}=\mathrm{resize}\!\Big(\mathrm{Conv}_{1\times 1}\big(\mathrm{FiLM}(\text{Res}(\mathbf{I}^{\text{src}}),c_{\text{src}})\big),\\(H,W)\Big)
\in\mathbb{R}^{B\times C \times H\times W},
\quad \text{src}\in\{\text{sat},\text{r-sd}\}.
\end{aligned}
\end{equation}
Then, we flatten feature maps to BEV tokens using $\mathrm{FlattenHW}(\cdot)$: $\mathbf{T}^{\text{src}}_{\text{ras}}=\mathrm{FlattenHW}(\mathbf{F}^{\text{src}}_{\text{ras}})\in\mathbb{R}^{B\times (HW)\times C}.$

\textbf{$\mathrm{SE}(2)$ Micro-alignment.}
We keep the same objective and regularize as in Sec.~\ref{Sec: vector encoder} but regress the pose from raster features plus the BEV reference instead of polylines to predict \((\Delta{x},\Delta{y},\Delta{\theta})\). With meters-per-pixel $(m_x^{\text{src}},m_y^{\text{src}})$, we form the normalized affine for \texttt{grid\_sample}:
\begin{equation}
t_x=\frac{2}{W-1}\frac{\Delta{x}}{m_x},\qquad\\
t_y=\frac{2}{H-1}\frac{\Delta{y}}{m_y},\qquad\\
\Theta=\begin{bmatrix}\cos \Delta{\theta} & -\sin \Delta{\theta} & t_x\\ \sin \Delta{\theta} & \cos \Delta{\theta} & t_y\end{bmatrix}.
\label{eq:affine}
\end{equation}
The aligned prior feature and tokens are
\begin{equation}
\widetilde{\mathbf{F}}^{\text{src}}_{\text{ras}}=\mathrm{grid\_sample}\big(\mathbf{F}^{\text{src}}_{\text{ras}},\,{G}(\Theta)\big),\qquad\\
\widetilde{\mathbf{T}}^{\text{src}}_{\text{ras}}=\mathrm{FlattenHW}\!\left(\widetilde{\mathbf{F}}^{\text{src}}_{\text{ras}}\right).
\label{eq:warp}
\end{equation}

\textbf{Presence-normalized Channel-wise Gated Fusion.}
We reuse the per-channel, presence-normalized softmax gates of Eq.~\eqref{eq:channelwise_gates} to  weight the two raster streams after $\mathrm{SE}(2)$:
\begin{equation}
\bar{\mathbf{Z}}
=
g_{\text{sat}}\odot \widetilde{\mathbf{T}}^{\text{sat}}_{\text{ras}}
+
g_{\text{r-sd}}\odot \widetilde{\mathbf{T}}^{\text{r-sd}}_{\text{ras}}
\;\in\;\mathbb{R}^{B\times(HW)\times C},
\label{eq:Z_raster_tokens}
\end{equation}
yielding fused raster prior \(\bar{\mathbf{Z}}\).

\subsection{Four Map Priors Residual Fusion}
\label{sec: four priors fusion}
We fuse the four priors in a vector-first, raster-second sequence that mirrors their roles: vectors provide precise geometry and topology; rasters supply dense appearance. 

\textbf{Vector stage (geometry first):} As Sec.~\ref{Sec: vector encoder} mentioned, HD/SD vector priors are encoded into polyline tokens with confidences. BEV queries $\mathbf{X}$ attend to each source separately via dual cross-attention (Eq.~\eqref{eq: cross attention}). Then we apply a channel-wise gate fusion, yielding $\bar{\mathbf{Y}} \in\mathbb{R}^{B\times(HW)\times C}$ (Fig.~\ref{fig:method} (a)).

\textbf{Raster stage (dense refinement):}
Unlike the vector path, we \textbf{do not} use BEV$\leftrightarrow$ prior cross-attention here: raster priors and the online BEV are both \textbf{image-like, pixel-aligned} on the BEV lattice, so a zero-initialized fusion preserves spatial locality and avoids the length-imbalance issues of attention over dense grids (Fig.~\ref{fig:method} (c)):
\begin{equation}
    \mathbf{X}_{\mathrm{UMPE}}=\mathrm{LN}(\bar{\mathbf{Y}})+\alpha\,\mathbf{W}\,\mathrm{LN}(\bar{\mathbf{Z}}),  
\end{equation}
where $\mathbf{W}\in\mathbb{R}^{C\times C}$ is a linear layer initialized to zero, $\alpha$ is a scalar scheduled from $0$ to $\sim 0.6$. LayerNorm aligns token scales to avoid one source dominating the fusion. The zero-initialized residual ensures the network starts from the BEV baseline and learns to add prior information only where it improves the mapping objective.
The final fused representation $\mathbf{X}_{\mathrm{UMPE}} \in\mathbb{R}^{B\times(HW)\times C}$ is then fed to the task-specific decoder for online mapping or end-to-end planning.
\section{Experiment}
We evaluate \emph{UMPE} under three hypotheses: (H1) \textbf{Mapping generality}—\emph{UMPE} yields consistent gains across datasets and baselines (Sec.~\ref{sec: online mapping results}); (H2) \textbf{Planning benefit}—\emph{UMPE} improves end-to-end planning (Sec.~\ref{sec: planning results}); (H3) \textbf{Module effectiveness \& any-subset robustness}—each design choice in \emph{UMPE} contributes measurably, and \emph{UMPE} handles arbitrary prior subsets (Sec.~\ref{sec: ablation study}).

\begin{table*}[t]
  \centering
  \small
  \caption{\textbf{Mapping results} on nuScenes validation dataset. Priors: VP--Vectorized HD/SD map Priors; RP--Rasterized SD map/satellite imagery Priors. Backbone: R--ResNet; T--Transformer. FPS is measured on a single RTX 3090. Since nuScenes lacks bulit-in HD maps, we follow~\cite{sun2023mind} to create an HD map by retaining only road boundaries and removing pedestrian crossings and dividers.}
  \renewcommand{\arraystretch}{0.8}
  \resizebox{\linewidth}{!}{
  \begin{tabular}{l|cc|cc|cccc|cc}
    \toprule
    \textbf{Method} & VP & RP & Backbone & Epoch &
    $\mathbf{AP}_{\text{ped}}$ &
    $\mathbf{AP}_{\text{div}}$ &
    $\mathbf{AP}_{\text{bou}}$ &
    \textbf{mAP} & \#Param.(M) & FPS $\uparrow$\\
    \midrule
    VectorMapNet~\cite{liu2023vectormapnet} & & & R50 & 110 & 42.5 & 51.4 & 44.1 & 46.0 & 36.7 & 16.17\\
    MapTR~\cite{maptr2022} & & & R50 &24  & 46.3 & 51.5 & 53.1 & 50.3 & 36.3 & 15.10\\
    StreamMapNet~\cite{Streammapnet2024} & & & R50 & 24 & 64.1 & 58.2 & 59.4 & 60.6 & 57.7 & 10.16\\
    MGMap~\cite{liu2024mgmap}  & &  & R50 & 30 & 61.8 & 65.0 & 67.8 & 64.8 & 55.9 & 11.60\\
\cmidrule(lr){1-11}
\morecmidrules
\cmidrule(lr){1-11}
    MapTRv2~\cite{Maptrv2} & & & R50 & 24 & 59.8 & 62.4 & 62.4 & 61.5 & 40.3  & 11.57 \\
     + SMERF~\cite{smerf2024}  & \cmark &  & R50\&T & 24 & 60.4 & 63.4 & 63.2 & 62.3 {\scriptsize\textcolor{green!55!black}{(+0.8)}} & 48.6 & 10.32\\
     + PriorDrive~\cite{unified_vector_prior}  & \cmark &  & R50\&T & 24 & 61.5 & 65.6 & 68.3 & 65.1 {\scriptsize\textcolor{green!55!black}{(+3.6)}} & 43.4 & --\\
     + SATP~\cite{SATP2025CVPR}  & \cmark &  & R50\&T & 24 & -- & -- & -- & 61.9  {\scriptsize\textcolor{green!55!black}{(+0.4)}} & 40.6 & --\\
      \rowcolor{gray!10}
    + Vector Encoder (Ours)  & \cmark &  & R50\&T & 24 & 63.0 & 68.6 & 68.8 & 66.8 {\scriptsize\textcolor{green!55!black}{(+5.3)}} & 43.5 & 11.37\\
\midrule
    + SatforHDMap~\cite{SatforHDmap_2024}  & & \cmark & R50\&R18 & 24  & 63.6 & 63.1 & 64.4 & 63.7  {\scriptsize\textcolor{green!55!black}{(+2.2)}}  & 58.0 & 10.32 \\
        \rowcolor{gray!10}
    + Raster Encoder (Ours)  & & \cmark & R50\&R18 & 24  & 65.7 & 66.1  & 68.2 & 66.7 {\scriptsize\textcolor{green!55!black}{(+5.2)}} & 51.8 & 11.36 \\
        \rowcolor{gray!10}
\midrule 
        \rowcolor{gray!20}
    + UMPE (Ours) & \cmark & \cmark & R50\&T\&R18 & 24 & 66.6 & 67.2 & 68.2 & 67.4 
 {\scriptsize\textcolor{green!55!black}{(+5.9)}} & 54.8 & 10.98 \\
\cmidrule(lr){1-11}
\morecmidrules
\cmidrule(lr){1-11}
 MapQR~\cite{liu2024mapQR}  & &  & R50 & 24 & 63.4 & 68.0 & 67.7 & 66.4  & 125.0 & 7.72\\
         \rowcolor{gray!10}
 + Vector Encoder (Ours)  &\cmark &  & R50 & 24 & 66.8 & 73.6 & 71.4 & 70.6 
 {\scriptsize\textcolor{green!55!black}{(+4.2)}} & 128.3  &  7.26\\
   \rowcolor{gray!10}
 + Raster Encoder (Ours)  & &\cmark  & R50 & 24 & 68.7 & 73.4 & 72.2 & 71.4 
 {\scriptsize\textcolor{green!55!black}{(+5.0)}} & 136.6 & 7.35 \\
         \rowcolor{gray!20}
+ UMPE (Ours)  &\cmark & \cmark & R50 & 24 & \textbf{69.0} & \textbf{73.8} & \textbf{72.3} & \textbf{71.7} 
 {\scriptsize\textcolor{green!55!black}{(+5.3)}} & 139.5 & 7.07 \\
    \bottomrule
  \end{tabular}
  }
  \label{tab:nuscnees_map_comparison}
\end{table*}

\begin{figure*}
    \centering
    \includegraphics[width=0.9\linewidth]{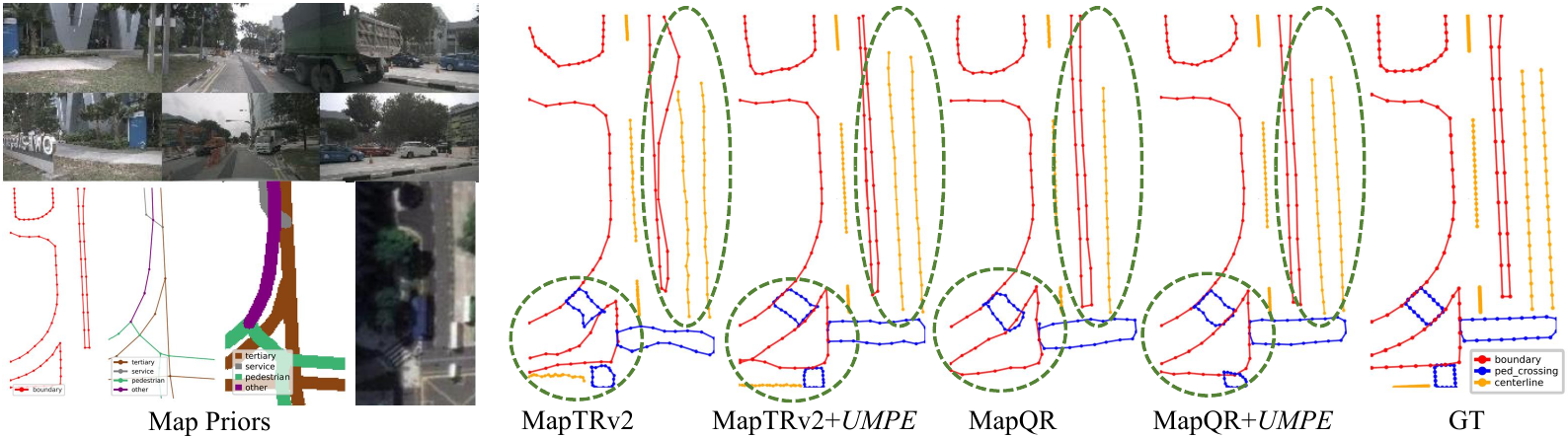}
    \caption{\textbf{Online mapping visualization on nuScenes}. Adding \emph{UMPE} to both MapTRv2~\cite{Maptrv2} and MapQR~\cite{liu2024mapQR} produces more accurate maps, especially in the green-highlighted regions: baselines show broken pedestrian crossings, kinked boundaries and missing dividers; \emph{UMPE} straightens, restores them.} 
    \label{fig:mapping_vis}
    \vspace{-0.4cm}
\end{figure*}

\subsection{Experimental Setup}

\textbf{Datasets and Metrics.}
We evaluate online HD map construction on nuScenes~\cite{2020nuscenes} and Argoverse2~\cite{wilson2023argoverse} using their official splits. For each frame, we extract four priors in an ego-centric BEV crop of $60{\rm m}\!\times\!30{\rm m}$: vectorized HD map, vectorized SD map, satellite imagery, and rasterized SD map. Following standard protocols for vectorized mapping, we report \textbf{mAP} computed from average precision over Chamfer distance thresholds $\tau\!\in\!\{0.5,1.0,1.5\}{\rm m}$ between predicted vectors and ground-truth map elements.

For end-to-end autonomous driving, we evaluate on nuScenes~\cite{2020nuscenes}. The same four priors are extracted per frame and injected into the policy via our unified map prior encoder. We report two standard metrics: \textbf{L2 error}—the mean Euclidean distance between the planned and ground-truth ego trajectories and \textbf{collision rate}—the frequency of rollouts where the ego trajectory  collides with other agents.

\textbf{Implementation Details.}
We train \emph{UMPE} from scratch with a two-stage curriculum and SourceDropout. We randomly drop one source with a 0.3 probability in each encoder. Stage 1: we optimize the full model with separate parameter groups—higher LR for the prior branches and a lower LR for the BEV encoder and map decoder. The residual scales $\alpha$ are linearly ramped from 0 to 0.2. Stage 2: we reduce LRs and relax $\alpha$ to 0.6, keeping fusion projections zero-initialized so the residual paths remain “do-no-harm” early. For mapping, we train 24 epochs on nuScenes and 6 epochs on Argoverse 2 using 4×RTX 3090; for planning, we train 60 epochs on nuScenes using 8×A800. 

\begin{figure*}
    \centering
    \setlength\abovecaptionskip{-2pt}
    \includegraphics[width=0.9\linewidth]{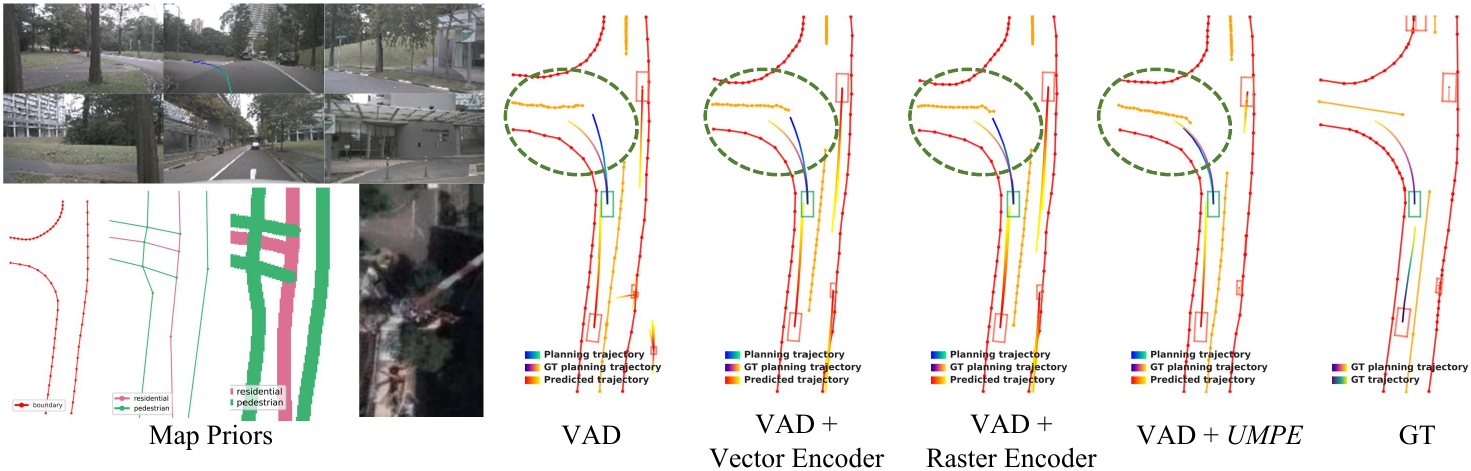}
    \caption{\textbf{End-to-end Planning visualization on nuScenes.} The ego vehicle is turning left. VAD without priors drifts toward the oncoming lane; adding the vector encoder or raster encoder improves lane adherence but leaves lateral error, while VAD+\emph{UMPE} produces a trajectory tightly overleaps the GT.}
    \label{fig:planning_vis}
    \vspace{-0.3cm}
\end{figure*}

\subsection{Online Mapping Results}
\label{sec: online mapping results}

\textbf{Quantitative Results.}
On the NueScens~\cite{2020nuscenes} dataset, we adopt MapTRv2~\cite{Maptrv2} as the baseline and compare against representative methods that inject either vector priors or raster priors (Tab.~\ref{tab:nuscnees_map_comparison}). Our \emph{UMPE} consistently surpasses others when using a single prior encoder and achieves the highest mAP when both prior encoders are enabled. Despite the accuracy gains, runtime remains comparable to prior work. To assess scalability, we also plug \emph{UMPE} into the stronger MapQR~\cite{liu2024mapQR} baseline and observe consistent improvements, indicating that our modules are model-agnostic and transferable across backbones. On Argoverse~2~\cite{wilson2023argoverse}, \emph{UMPE} again improves over the baseline and clearly outperforms other prior-fusion methods, confirming its robustness across datasets and scalable fusion (Tab.~\ref{tab:argo2_map_comparison}). 

Per-class trends are consistent with prior semantics. The vector encoder chiefly improves divider and boundary AP since its accurate lane geometry. The raster encoder contributes most on pedestrian crossing AP, as satellite input provides top-down texture for crosswalk patterns. 

\textbf{Qualitative Results.} As shown in Fig.~\ref{fig:mapping_vis}, \emph{UMPE} regularizes the baseline geometry: fragmented pedestrian crossings become closed and well-shaped, lane boundaries and dividers straighten and align with the road layout. These corrections appear consistently across backbones, bringing the predictions noticeably closer to the ground truth.  

\begin{table}[t]
  \centering
  \small
    \caption{\textbf{Mapping results} on Argoverse~2~\cite{wilson2023argoverse} validation dataset. }
  \renewcommand{\arraystretch}{0.8}
  \resizebox{\linewidth}{!}{
  \begin{tabular}{l|cccc}
    \toprule
    \textbf{Method} &
    $\mathbf{AP}_{\text{ped}}$ &
    $\mathbf{AP}_{\text{div}}$ &
    $\mathbf{AP}_{\text{bou}}$ &
    \textbf{mAP} \\
    \midrule
    VectorMapNet~\cite{liu2023vectormapnet}  & 35.6 & 34.9 & 37.8 & 36.1 \\
    MapTR~\cite{maptr2022} & 48.1 & 50.4 & 55.0 & 51.1 \\
    StreamMapNet~\cite{Streammapnet2024} & 56.9 & 55.9 & 61.4 & 58.1\\
    MGMap~\cite{liu2024mgmap}  & 52.8 & 67.5 & 68.1 & 62.8 \\
    MapQR~\cite{liu2024mapQR}  & 60.1 & 71.2 & 66.2 & 65.9 \\
    \midrule
    MapTRv2~\cite{Maptrv2} & 60.7 & 68.9 & 64.5 & 64.7 \\
    + SMERF~\cite{smerf2024}  & 60.5 & 69.1 & 67.8 & 65.8  {\scriptsize\textcolor{green!55!black}{(+1.1)}}\\
    + SDTagNet~\cite{sdtagnet2025}  & 62.1 & 66.1 & 70.7 & 66.3  {\scriptsize\textcolor{green!55!black}{(+1.6)}}\\
    \rowcolor{gray!10}
    + Vector Encoder (Ours)  & 62.4 & 67.7 & \textbf{70.9} & 67.0 {\scriptsize\textcolor{green!55!black}{(+2.3)}}\\
     \rowcolor{gray!10}
    + Raster Encoder (Ours)  & 63.7 & \textbf{72.6} & 67.4 & 67.9 {\scriptsize\textcolor{green!55!black}{(+3.2)}}\\
    \rowcolor{gray!20}
    + UMPE (Ours) & \textbf{63.8} & \textbf{72.6} & 70.0 & \textbf{68.8} {\scriptsize\textcolor{green!55!black}{(+4.1)}}\\
    \bottomrule
  \end{tabular}
  }
  \vspace{-0.4cm}
  \label{tab:argo2_map_comparison}
\end{table}

\vspace{-10pt}
\subsection{End-to End Autonomous Driving Results}
\label{sec: planning results}

\textbf{Quantitative Results.}
We have shown that incorporating map priors into a unified encoder significantly improves mapping performance. A natural next step is to ask whether these improvements in perception and map understanding can transfer to the planning domain. We therefore plug the \emph{UMPE} into VAD~\cite{jiang2023vad} as an auxiliary prior-fusion branch that augments the BEV feature before the planning decoder. We compare \emph{UMPE} against three representative prior-injection methods~\cite{SATP2025CVPR, pmapnet2024, unified_vector_prior} with the same VAD backbone. Tab.~\ref{Tab:planning_results} shows that our vector encoder already yields a large $\mathbf{L}_2$ drop, while the raster encoder also helps. Combining both in \emph{UMPE} gives the strongest improvement ($\mathbf{L}_2$ avg -0.30m vs. VAD and Collision Rate -0.10\% vs. VAD), outperforming all prior-injection methods. The effect is mainly because vector priors provide metrically accurate geometry selected via confidence-biased cross-attention, while raster priors add dense drivable context; the two are complementary.

\textbf{Qualitative Results.} Fig.~\ref{fig:planning_vis} corroborates the quantitative results. We attribute these gains to stronger BEV perception and mapping. Once these map priors are fused, the planner “sees” a more structured map, leading to lower trajectory error and collisions. That is better mapping, better planning.

\begin{table}[t]
  \centering
  \small
   \caption{\textbf{Planning results} on the nuScenes~\cite{2020nuscenes} validation dataset. 
  }
    \renewcommand{\arraystretch}{1.0}
  \resizebox{\linewidth}{!}{
  \begin{tabular}{l|ccc>{\columncolor{gray!8}}c|ccc>{\columncolor{gray!8}}c}
    \toprule
    \multirow{2}{*}{\textbf{Method}} &
    \multicolumn{4}{c|}{\textbf{L2 (m)} $\downarrow$} &
    \multicolumn{4}{c}{\textbf{Col. Rate (\%)} $\downarrow$} \\
    \cmidrule(lr){2-5}\cmidrule(lr){6-9}
     & 1s & 2s & 3s & Avg. & 1s & 2s & 3s & Avg. \\
    \midrule
    
    UniAD~\cite{UniAD} & 0.45 & 0.70 & 1.04 & 0.73 & 0.62 & 0.58 & 0.63 & 0.61 \\
    GenAD~\cite{2024genad} & 0.28 & 0.49 & 0.78 & 0.52 & 0.08 & 0.14 & 0.34 & 0.19 \\
    SparseDrive~\cite{sun2024sparsedrive} & 0.29 & 0.58 & 0.96 & 0.61 & \textbf{0.01} & \textbf{0.05} & 0.18 & \textbf{0.08} \\
    BEVPlanner~\cite{bevplanner2024} & 0.27 & 0.54 & 0.90 & 0.57 & -- & -- & -- & -- \\
    Epona~\cite{2025epona} & 0.61 & 1.17 & 1.98 & 1.25 & \textbf{0.01} & 0.22 & 0.85 & 0.36 \\
    MomAD~\cite{2025MomAD} & 0.31 & 0.57 & 0.91 & 0.60 & \textbf{0.01} & \textbf{0.05} & 0.22 & 0.09 \\
    BridgeAD~\cite{2025bridgeAD} & 0.29 & 0.57 & 0.92 & 0.59 & \textbf{0.01} & \textbf{0.05} & 0.22 & 0.09 \\
    DiffusionDrive~\cite{2025diffusiondrive} & 0.27 & 0.54 & 0.90 & 0.57 & 0.03 & \textbf{0.05} & \textbf{0.16} & \textbf{0.08} \\
    \midrule
    VAD~\cite{jiang2023vad} & 0.41 & 0.70 & 1.05 & 0.72 & 0.07 & 0.17 & 0.41 & 0.22 \\
    + SATP~\cite{SATP2025CVPR} & 0.39 & 0.62 & 0.94 & 0.65  {\scriptsize\textcolor{green!55!black}{(-0.07)}}& 0.14 & 0.21 & 0.42 & 0.26 {\scriptsize\textcolor{red!55!black}{(+0.04)}}\\
    + PmapNet~\cite{pmapnet2024} & 0.35 & 0.65 & 1.05 & 0.68  {\scriptsize\textcolor{green!55!black}{(-0.04)}}& 0.37 & 0.54 & 1.06 & 0.66 {\scriptsize\textcolor{red!55!black}{(+0.44)}}\\
    + PriorDrive~\cite{unified_vector_prior} & 0.35 & 0.68 & 1.16 & 0.73  {\scriptsize\textcolor{red!55!black}{(+0.01)}}& 0.26 & 0.46 & 1.12 & 0.61 {\scriptsize\textcolor{red!55!black}{(+0.39)}}\\
    \rowcolor{gray!15}
     + Vector Enc (Ours) & 0.25 & 0.45 & 0.74 & 0.48 {\scriptsize\textcolor{green!55!black}{(-0.24)}}& 0.09 & 0.11 & 0.31 &  0.17 {\scriptsize\textcolor{green!55!black}{(-0.05)}}\\
     \rowcolor{gray!15}
    + Raster Enc (Ours) & 0.28 & 0.53 & 0.83 & 0.55 {\scriptsize\textcolor{green!55!black}{(-0.17)}}& 0.09 & 0.13 & 0.35 & 0.19 {\scriptsize\textcolor{green!55!black}{(-0.03)}}\\
    \rowcolor{gray!20}
    + UMPE (Ours) & \textbf{0.21} & \textbf{0.39} & \textbf{0.65} & \textbf{0.42} {\scriptsize\textcolor{green!55!black}{(-0.30)}}& 0.03 & 0.08 & 0.25 & 0.12 {\scriptsize\textcolor{green!55!black}{(-0.10)}}\\
    \bottomrule
  \end{tabular}
  }

  \label{Tab:planning_results}
  \vspace{-0.5cm}
\end{table}

\subsection{Ablation Study}
\label{sec: ablation study}
\textbf{Vector Encoder Modules.}
To verify the effectiveness of each component in our vector encoder, we conduct a step-by-step ablation on the MapTRv2 trained for 24 epochs (Tab.~\ref{Tab:vector encoder ablation}). First, for tokenization, \textbf{Sine-PE} outperforms raw (x,y)→MLP (2 vs. 3), indicating that multi-frequency point encoding better preserves fine lane geometry. With tokenization fixed, switching the fusion strategy from concatenation + single cross-attention to \textbf{dual cross-attention} yields a clear gain (2 vs. 4), showing that separating HD/SD avoids length imbalance in the softmax. Adding the \textbf{confidence bias} inside attention brings further improvement (4 vs. 5) by down-weighting uncertain polylines. Incorporating presence-normalized \textbf{gating} adds another gain (5 vs. 6). Finally, appending the \textbf{SE(2) pre-alignment} on top performs best (6 vs. 7) by removing small pose drift before tokenization.

\begin{table*}[t]
\centering
\small
\caption{Ablations on \textbf{vector-prior fusion}.
\textbf{MLP}: replace Sine-PE with raw $(x,y)$ to a polyline MLP. \textbf{Single-Attn}: concat HD/SD then a single cross-attention. \textcolor{green!55!black}{(+$\Delta$)} denotes the absolute gain over the baseline~\cite{Maptrv2}. All others Modules are defined in Sec.~\ref{Sec: vector encoder}.}
\renewcommand{\arraystretch}{0.5}
\resizebox{\linewidth}{!}{
\begin{tabular}{
                l| c c | c c c| c| c 
                | c c c c | c}
\toprule
\multirow{2}{*}{ID} & \multicolumn{7}{c|}{Vector Encoder Modules} &
\multicolumn{4}{c|}{AP $\uparrow$} & \multirow{2}{*}{\#Param.(M)}\\
\cmidrule(lr){2-8}\cmidrule(lr){9-12}  
 &  Sine-PE & MLP  &  Single-Attn & Dual-Attn & ConfBias  & Gating & SE(2)& 
ped. & div. & bou. & mean & \\
\midrule
1&  &  &      &  &   &      &   & 59.8 & 62.4 & 62.4 & 61.5  & 40.3\\
2 & \cmark &  & \cmark     &     &      &  &  & 61.2 & 65.4 & 64.1 & 63.6 {\scriptsize \textcolor{green!55!black}{(+2.1)}} & 43.2\\
3&  & \cmark   & \cmark &  &   &      &  & 61.1 & 63.5 & 62.5 & 62.3 {\scriptsize\textcolor{green!55!black}{(+0.8)}} & 45.9\\
4 & \cmark &        &  & \cmark &   &       &  & 61.7 & 68.0 & 64.9 & 64.9 {\scriptsize\textcolor{green!55!black}{(+3.4)}} & 43.3\\
5 & \cmark &      &  &  \cmark      & \cmark  &      &   & 62.1 & 68.3 & 65.8 & 65.4 {\scriptsize\textcolor{green!55!black}{(+3.9)}}& 43.3\\
6&  \cmark &  &      & \cmark & \cmark   & \cmark     &  & 62.5 & 68.5 & 66.0 & 65.7 {\scriptsize\textcolor{green!55!black}{(+4.2)}} & 43.3\\
    \rowcolor{gray!20}
7 &  \cmark &  &      & \cmark & \cmark  &  \cmark    &  \cmark    & \textbf{63.0} & \textbf{68.6}  & \textbf{68.8} & \textbf{66.8} {\scriptsize\textcolor{green!55!black}{(+5.3)}}& 43.5\\
\bottomrule
\end{tabular}
}

\label{Tab:vector encoder ablation}
\vspace{-0.3cm}
\end{table*}

\textbf{Raster Encoder Modules.}
On MapTRv2, we ablate the raster encoder under the same backbone and schedule (Tab.~\ref{Tab:raster encoder ablation}). First, applying 
FiLM at \textbf{every} ResNet stage outperforms late-stage FiLM (2 vs. 3), indicating that distributed modulation better handles domain shift across layers. With SH-FiLM fixed, \textbf{feature-based gating} (Feagate) is clearly preferable to conditioning-vector gating (Congate)(4 vs. 5), showing that gates should depend on encoded raster evidence rather than metadata alone. The zero-initialized \textbf{residual} injection is the largest contributor to accuracy (2 vs. 6). Finally, adding \textbf{SE(2) micro-alignment} on the full setting polishes residual pose/scale mismatches (7 vs. 8). Overall, the complete raster path delivers +5.2 mAP over the baseline.

\begin{table}[t]
\centering
\small
\caption{Ablations on \textbf{Raster-prior fusion}. \textbf{FiLM placement}: apply FiLM at every stage of the ResNet backbone (\textbf{SH-FiLM}) or only at the last stage (\textbf{LY-FiLM}).
\textbf{Gating}: predicted either from the conditioning vector (\textbf{Congate}) or from raster features (\textbf{Feagate}).
Other modules are defined in Sec.~\ref{Sec: raster encoder}.}
\renewcommand{\arraystretch}{1.0}
\resizebox{\linewidth}{!}{
\begin{tabular}{
                l | c c | c c c| c 
                | c c c c}
\toprule

\multirow{2}{*}{ID} &  \multicolumn{6}{c|}{Raster Encoder Modules} &
\multicolumn{4}{c}{AP $\uparrow$} \\
\cmidrule(lr){2-7}\cmidrule(lr){8-11}
 &   SH-FiLM & LY-FiLM  & Congate  & Feagate & Residual & SE(2)&
ped. & div. & bou. & mean \\
\midrule
1 & &  &      &  &   &       & 59.8 & 62.4 & 62.4 & 61.5\\
2 & \cmark &  &      &  &   &       & 62.5 & 64.5 & 64.0 & 63.7 {\scriptsize\textcolor{green!55!black}{(+2.2)}}\\
3 &         & \cmark &      &  &    &  & 62.1 & 64.0 & 63.1 & 63.1 {\scriptsize\textcolor{green!55!black}{(+1.5)}}\\
4 & \cmark &        & \cmark &  &    &  & 59.9 & 62.8 & 62.9 & 61.8 {\scriptsize\textcolor{green!55!black}{(+0.3)}}\\
5 & \cmark &        &  & \cmark &   &  & 63.5 & 66.1 & 65.0 & 64.9 {\scriptsize\textcolor{green!55!black}{(+3.4)}}\\
6 & \cmark &        &  &  &  \cmark &  & 65.4 & 67.1 & 66.1 & 66.2 {\scriptsize\textcolor{green!55!black}{(+4.7)}}\\
7 & \cmark &        &  &  \cmark      & \cmark  &  &65.5 & 67.1 & 66.5 & 66.4 {\scriptsize\textcolor{green!55!black}{(+4.9)}}\\

    \rowcolor{gray!20}
8 &  \cmark &  &      & \cmark & \cmark  &  \cmark & \textbf{65.6} & \textbf{67.4}  & \textbf{66.9} & \textbf{66.7} {\scriptsize\textcolor{green!55!black}{(+5.2)}}\\
\bottomrule
\end{tabular}
}
\vspace{-0.3cm}
\label{Tab:raster encoder ablation}
\end{table}

\textbf{Fusion Order.} We test whether the order of fusing vector and raster priors affects the fusion performance. In \textbf{V$\rightarrow$R}, BEV tokens first absorb vector priors via dual cross-attention and then fuse raster priors through residual projection. In \textbf{R$\rightarrow$V}, BEV tokens are first updated by raster residual fusion before serving as queries for vector cross-attention. 
Tab.~\ref{tab:fusion_order} shows vector-first raster-second fusion order is better. 
Vector-first preserves clean BEV queries for cross-attention, whereas raster-first injects dense, potentially misaligned signals that degrade attention selectivity. The order also matches the inductive bias: vectors set the global geometric structure, and raster then provides local appearance refinements, yielding better convergence and higher AP.

\begin{table}[t]
  \centering
  \small
    \caption{Ablations on \textbf{fusion order} of vector and raster priors. }
  \renewcommand{\arraystretch}{0.8}
  \resizebox{0.8\linewidth}{!}{
  \begin{tabular}{l|cccc}
    \toprule
    Fusion Order &
    $\mathbf{AP}_{\text{ped}}$ &
    $\mathbf{AP}_{\text{div}}$ &
    $\mathbf{AP}_{\text{bou}}$ &
    \textbf{mAP} \\
    \midrule
    Raster$\rightarrow$Vector & 63.8 & 65.1 & 66.7 & 65.2 \\
    \rowcolor{gray!20}
    Vector$\rightarrow$Raster  & 66.6 & 67.2 & 68.2 & 67.4 \\
    \bottomrule
  \end{tabular}
  }
  \label{tab:fusion_order}
  \vspace{-0.5cm}
\end{table}

\textbf{Arbitrary Combinations of Map Priors.}
We assess source robustness and compositionality of \emph{UMPE} by evaluating subsets of the four priors on MapTRv2. Concretely, we compare (i) single-prior baselines trained with that source only to (ii) our final unified encoder trained with all priors but toggled at test time (no retraining) (Tab.~\ref{Tab:map priors ablation}).
This tests that our encoder can consume \textbf{any} available subset without retraining, in contrast to prior methods that assume a fixed prior modality or availability. Surprisingly, when we toggle to a single prior at test time, \emph{UMPE} outperforms models trained only on that single prior. We attribute this to two factors: (i) \textbf{Shared-target, multi-source co-training}. All priors are optimized together. The model learns complementary cues and avoids overfitting to any one modality's biases, so each branch is stronger even in isolation. (ii) \textbf{Robustness by design}. Zero-initialized residual fusion makes each branch modular, while SourceDropout exposes the network during training to missing prior scenarios. 

\begin{table}[t]
\centering
\small
\caption{\textbf{Effect of combining four map priors on mapping.} 
\textcolor{black}{\colorbox{green!12}{Green}} rows are \emph{single-prior} baselines (trained with that prior only).
All other rows use our \textbf{final unified encode}r trained with \emph{all} priors, with subsets toggled \emph{at test time only} (no retraining). }
\renewcommand{\arraystretch}{0.6}
\resizebox{\linewidth}{!}{
\begin{tabular}{
                c c | c c| c c c c}
\toprule

\multicolumn{4}{c|}{Map Priors} &
\multicolumn{4}{c}{AP $\uparrow$} \\
\cmidrule(lr){1-4}\cmidrule(lr){5-8}
  HD (vec) & SD (vec)  & Sat.Ima.   & SD (ras)&
ped. & div. & bou. & mean \\
\midrule
 &   &   &       & 59.8 & 62.4 & 62.4 & 61.5 \\
 \rowcolor{green!12}
\cmark &   &   &       & 59.7 & 62.5 & 66.5 & 62.9 {\scriptsize\textcolor{green!55!black}{(+1.4)}}\\
\cmark &   &   &       & 60.1 & 62.7 & 67.7 & 63.5 {\scriptsize\textcolor{green!55!black}{(+2.0)}}\\
 \rowcolor{green!12}
        &  \cmark   &    &  & 61.1 & 63.5 & 62.5 & 62.3 {\scriptsize\textcolor{green!55!black}{(+0.8)}}\\
        &  \cmark   &   &   & 62.3 & 64.3 & 63.6 & 63.4 {\scriptsize\textcolor{green!55!black}{(+1.9)}}\\        
\cmark &  \cmark      &    &  & 63.0 & 68.6 & 68.8 & 66.8 {\scriptsize\textcolor{green!55!black}{(+5.3)}} \\
\midrule
 \rowcolor{green!12}
 &       & \cmark  &  & 63.7 & 65.4 & 64.5 & 64.5 {\scriptsize\textcolor{green!55!black}{(+3.0)}} \\
 &       & \cmark  &  & 63.4 & 65.4 & 65.9 & 64.9 {\scriptsize\textcolor{green!55!black}{(+3.4)}} \\ 
  \rowcolor{green!12}
 &         &   & \cmark & 62.1 & 65.1 & 64.3 & 63.8 {\scriptsize\textcolor{green!55!black}{(+2.3)}} \\
  &         &   & \cmark & 63.9 & 65.9 & 65.9 & 65.3 {\scriptsize\textcolor{green!55!black}{(+3.8)}} \\
   &         &  \cmark & \cmark & 65.6 & 67.4  & 66.9 & 66.7 {\scriptsize\textcolor{green!55!black}{(+5.2)}}  \\
   \midrule
 &  \cmark      & \cmark  &  & 66.1 & 67.1 & 67.2 & 66.8 {\scriptsize\textcolor{green!55!black}{(+5.3)}} \\
  &  \cmark      &   & \cmark & 65.4 & 66.8 & 66.4 & 66.2 {\scriptsize\textcolor{green!55!black}{(+4.7)}}\\
    &  \cmark      & \cmark  & \cmark & 66.3 & 66.8 & 67.6 & 66.9 {\scriptsize\textcolor{green!55!black}{(+5.4)}}\\
 \midrule
    \rowcolor{gray!20}
 \cmark &  \cmark & \cmark  &  \cmark & \textbf{66.6} & \textbf{67.2} & \textbf{68.2} & \textbf{67.4} 
 {\scriptsize\textcolor{green!55!black}{(+5.9)}} \\
\bottomrule
\end{tabular}
}
\label{Tab:map priors ablation}
\vspace{-0.4cm}
\end{table}
\section{Conclusion}

We presented \emph{UMPE}, a unified, alignment-aware encoder that accepts any subset of four complementary map priors and fuses them with BEV features for online mapping and end-to-end planning. Across nuScenes and Argoverse 2, \emph{UMPE} delivers consistent mAP gains on strong backbones and, when plugged into VAD, substantially reduces trajectory $\mathbf{L}_2$ and collision rate, surpassing recent prior-injection baselines. The encoder is compositional and robust to missing sources, enabling test-time toggling without retraining. Future work will prioritize investigating the role of map priors in closed-loop, end-to-end autonomous driving.

{
\small
\bibliographystyle{ieeetr}
\bibliography{main}
}

\end{document}